\documentclass[12pt]{article} 

\usepackage{amssymb}
\usepackage{amsmath}
\usepackage{graphicx}
\usepackage{subfigure}
\usepackage{makeidx}
\usepackage{multicol}
\usepackage[export]{adjustbox}
\usepackage[justification=centering]{caption}
\usepackage{chngcntr}
\counterwithout{figure}{section}
\usepackage{float}
\usepackage{booktabs} 
\usepackage{xcolor}
\def\B#1{\textcolor{black}{#1}}

\usepackage{fancyhdr}
\let\origtitle\title 
\renewcommand{\title}[1]{\lfoot{\textit{#1}}\origtitle{\textbf{#1}}}
\renewcommand{\sectionmark}[1]{\markboth {}{}}

\date{}

\title{Diffusion Model-based Activity Completion for AI Motion Capture from Videos}

\begin{document}
\maketitle

\author{Gao Huayu \footnote{gao.huayu934@mail.kyutech.jp},  Huang Tengjiu \footnote{huang.tengjiu275@mail.kyutech.jp}, Ye Xiaolong \footnote{ye.xiaolong655@mail.kyutech.jp}, Tsuyoshi Okita* \footnote{tsuyoshi@ai.kyutech.ac.jp}}\\
\thanks{\begin{center} 
    Kyushu Institute of Technology \end{center}}

\abstract{

AI-based motion capture is an emerging technology that offers a cost-effective alternative to traditional motion capture systems. However, current AI motion capture methods rely entirely on observed video sequences, similar to conventional motion capture. This means that all human actions must be predefined, and movements outside the observed sequences are not possible.
To address this limitation, we aim to apply AI motion capture to virtual humans, where flexible actions beyond the observed sequences are required. We assume that while many action fragments exist in the training data, the transitions between them may be missing. To bridge these gaps, we propose a diffusion-model-based action completion technique that generates complementary human motion sequences, ensuring smooth and continuous movements.
By introducing a gate module and a position-time embedding module, our approach achieves competitive results on the Human3.6M dataset. Our experimental results show that (1) MDC-Net outperforms existing methods in ADE, FDE, and MMADE but is slightly less accurate in MMFDE, (2) MDC-Net has a smaller model size (16.84M) compared to HumanMAC (28.40M), and (3) MDC-Net generates more natural and coherent motion sequences. Additionally, we propose a method for extracting sensor data, including acceleration and angular velocity, from human motion sequences.
}

\section{Introduction}


\B{Motion capture (MoCap), also known as motion tracking, involves recording and processing the movements of humans or objects. It has widespread applications across various fields, including the military, entertainment, sports, medical applications, computer vision, and robotics \cite{tang2024screen}.
Traditional motion capture systems, however, rely on expensive hardware setups, such as complex optical cameras and motion sensors, which limit their scalability and accessibility. Additionally, these systems often face challenges with real-time data processing and performance in uncontrolled environments.
Computer vision addresses some of these limitations by using cameras and AI algorithms to capture, track, and analyze motion, enabling automatic motion recognition \cite{zhang2022using}. Techniques such as pose estimation \cite{cao2017realtime, moon2019posefix, sun2019deep, varol2017learning, wang2023freeman,Adachi2021,Mamun2024,Zolfaghari2024,Phan2024} and mesh estimation \cite{ma20233d,moon2022accurate} have significantly advanced motion tracking. However, these methods still rely entirely on observed video sequences, similar to traditional motion capture, restricting their ability to generate new, unseen motions.
Human motion generation offers a solution to this limitation. Recent advances in generative models have enabled the efficient and cost-effective synthesis of diverse human motion sequences, expanding the possibilities for AI-driven motion capture.}

With advancements in deep learning and GPU technology, human motion generation has rapidly evolved. Liu et al. \cite{liu2021aggregated} employed Generative Adversarial Networks (GANs) \cite{goodfellow2020generative} to generate new motion sequences from historical pose data. In 3D human motion generation, Xu et al. \cite{xu2023actformer} introduced ActFormer, a GAN-based Transformer \cite{vaswani2017attention} framework. ActFormer leverages a Transformer-based architecture to generate human motion sequences from an implicit vector and a given action class label.
Recent studies have utilized Denoising Diffusion Probabilistic Models (DDPMs) for human motion generation, focusing on control signals such as textual descriptions \cite{zhang2022motiondiffuse, liu2023plan, guo2024momask}, video \cite{ho2022video, skorokhodov2022stylegan}, images \cite{karras2021alias}, and 3D objects \cite{zhu2023human}. However, these approaches typically generate only short, disconnected motion sequences.
\B{To address this limitation, we propose a motion completion algorithm that enables the seamless concatenation of two human motion sequences of arbitrary length.} Our goal is to generate an intermediate sequence that smoothly connects two distinct motion segments, forming a coherent and continuous motion while also extracting IMU data.
\B{For example, consider two human motion sequences,
$ H1 \{ X_1, X_2, ... , X_n \} $ and $H2\left \{ Y_1, Y_2, ... , Y_n \right \}$, each consisting of N frames. By generating an intermediate sequence $ P \{ P_1, P_2, ... , \allowbreak P_m\}$, 
we form a new, smoothly transitioned motion sequence:
$H1$ and $H2$ to form a new sequence 
$ S\{X_1, X_2, ... , X_n, P_1, P_2, ... , \allowbreak P_m, Y_1, Y_2, ... , Y_k \}$.
Previous studies \cite{wan2023diffusionphase, holden2017phase, starke2022deepphase, zhang2018mode} have demonstrated that human movements exhibit periodicity. 
Leveraging this characteristic, the transition between $H1$ and $H2$
can be predicted using the final frames of $H1$ and the initial frames of $H2$. To achieve this, we employ a masking completion technique \cite{chen2023humanmac}, which facilitates the efficient extraction and integration of transition sequences.}


\begin{figure}[H]
\centering
\includegraphics[width=1.0\linewidth]{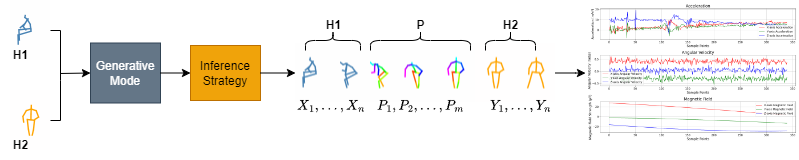}
\caption{Human Motion Completion. 
$H1$ and $H2$ are two human motions that can either be different or the same. Using a generative model and inference, we produce an intermediate motion sequence, P, to connect and complete these two motions.}
\label{fig1}
\end{figure}

\B{FlowMDM \cite{barquero2024seamless} and HumanMac \cite{chen2023humanmac} are diffusion model-based approaches designed to handle variations in motion types and transitions. In HumanMac, motion switching is constrained to a fixed length of 125 frames. Our method overcomes this limitation by allowing motion transitions to occur at any point within a sequence and generating motion completions of arbitrary length. For example, our approach enables the extension of a running motion, allowing a person to continue running indefinitely or transition smoothly into a sitting posture.
For our IMU data task, our process consists of four key steps:
\begin{itemize}
    \item  Generating the desired human motion sequences.
\item Predicting the 3D human body model.
\item Calculating the normal vectors of specific joints from the mesh model.
\item Inputting the normal vectors and XYZ coordinates into the MATLAB IMU module to simulate IMU data.
\end{itemize}
}
\B{To support the IMU data extraction process, we leverage existing frameworks, including SMPL \cite{loper2023smpl}, MotionBert \cite{motionbert}, and Neural Body \cite{peng2021neural}.}



In this paper, we propose a diffusion-based action completion framework to overcome the limitations of existing AI motion capture systems.

Our key contributions are as follows:
\begin{itemize}
\item \B{We introduce MDC-Net (Motion Diffusion Completion Network), which generates motion sequences to seamlessly connect two different human motions, creating longer and more continuous motion sequences.
\item HumanMAC introduced the concept of motion switching to generate transition sequences between independent actions. However, its generated sequences are relatively short (limited to 125 frames), and the model requires extensive training time.} To address these limitations, we designed a novel noise prediction network incorporating a gate module and a position-time embedding module. \item Our approach accepts motion sequences of any length as input and generates motion sequences of arbitrary length while maintaining a smaller model size and requiring less training time compared to HumanMAC.
\item \B{The extended motion sequences generated by our method provide richer and more flexible motion samples for virtual humans.}
\item \B{We propose a method to extract sensor data, including acceleration and angular velocity, from human motion sequences.}
\end{itemize}

\section{Related Literature}
In this section, we provide an overview of key technologies related to AI-based motion capture.

Human motion generation is widely used in film production, AR/VR, video games, robotics \cite{suzuki2022augmented, park2022metaverse}, and human-computer interaction due to its ability to accurately capture and replicate human movement.
\B{Beyond entertainment, AI-powered motion capture is also being utilized in sports analytics, medical rehabilitation, and metaverse-based virtual avatars. In sports, it helps track and analyze athletes' movements to optimize training regimens. In healthcare, AI-driven motion tracking assists in monitoring patient mobility, particularly in the rehabilitation of neurological disorders like Parkinson’s disease. Additionally, AI-powered MoCap enables virtual avatars to replicate human gestures in real time, enhancing digital interactions in the metaverse.}
\B{
Traditional motion capture systems, such as optical and IMU-based solutions, require either expensive multi-camera setups or wearable sensors. Optical MoCap provides high-precision tracking but is costly and environment-dependent, whereas IMU-based systems offer portability but suffer from sensor drift over time. In contrast, AI-driven motion capture leverages deep learning and computer vision to estimate human motion using simple RGB cameras or low-cost IMU sensors, making it more accessible and scalable for real-world applications.
AI motion capture primarily consists of deep learning-based visual methods and sensor-based AI computing methods. The former relies on RGB or RGB-D cameras for human pose estimation (HPE), with key techniques including 2D keypoint detection \cite{cao2017realtime, sun2019deep}, 3D motion reconstruction \cite{pavllo20193d}, and temporal optimization \cite{motionbert}. BoDiffusion \cite{bodiffusion} reconstructs full-body motion using only three tracking signals. It innovatively frames full-body tracking as a conditional sequence generation task and employs global joint positions and rotations as control signals, significantly improving the accuracy of lower-body motion predictions.
}

\B{Despite its advantages, AI motion capture still faces several challenges. One major issue is generalization, as deep learning models are typically trained on controlled datasets and often struggle to adapt to unseen environments. Additionally, occlusion remains a challenge—when certain body parts are blocked from the camera’s view, pose estimation accuracy can suffer. Lastly, real-time performance is critical for interactive applications such as VR and robotics, yet many high-precision AI MoCap models demand substantial computational resources. Addressing these challenges remains an active area of research in AI-driven motion analysis.}

\subsection{Diffusion Model}
Diffusion models have gained popularity as generative models due to their ability to produce high-quality samples. They have demonstrated remarkable success in image generation and have recently been adapted for other domains, including motion synthesis and human motion generation. The core concept of a diffusion model involves a process in which data is gradually transformed into noise over multiple steps and then reconstructed by reversing this process \cite{ho2020denoising, song2020score}. The forward process, which incrementally adds noise to the data, is described as:
\B{
\begin{equation}
q(\mathbf{x}_t | \mathbf{x}_{t-1}) = \mathcal{N}(\mathbf{x}_t; \sqrt{1 - \beta_t}\mathbf{x}_{t-1}, \beta_t \mathbf{I})
\end{equation}
}
\B{where $ x_t $ represents the noisy data at step $ t $, and $ \beta_t $  is a noise schedule that determines the amount of noise added at each step. 
The reverse process, which reconstructs the original data from the noise, is typically parameterized by a neural network,
\(\epsilon_\theta(\mathbf{x}_t, t)\), which predicts the noise at each step:}
\B{
\begin{equation}
p_{\theta}(\mathbf{x}_{t-1} | \mathbf{x}_t) = \mathcal{N}(\mathbf{x}_{t-1}; \mu_\theta(\mathbf{x}_t, t), \sigma_\theta(t) \mathbf{I})
\end{equation}
}
\B{where $\mu_\theta(\mathbf{x}_t, t)$ is the mean of the distribution predicted by the neural network, and $\sigma_\theta(t)$ is the variance. The model is trained to reverse the noise process by minimizing a loss function, typically a variant of the denoising score matching objective \cite{song2020score}. Diffusion models have been successfully applied to generate realistic human motions from noise, with some variations in models such as the score-based diffusion model \cite{ho2020denoising} improving the quality of generated sequences.} 
\B{
\subsection{Human Motion Generation and 3D Reconstruction}
 SMPL (Skinned Multi-Person Linear Model) \cite{loper2023smpl} is a parametric model used to generate 3D human body models. By linearly combining shape and pose parameters, it can produce human models with different body types and posesc\cite{zhu2024champ}. This model is widely used in computer vision, animation, and virtual reality, offering an efficient and adjustable way to generate and manipulate 3D human data. Zhang et al. \cite{zhang2024dstformer}. proposed a framework comprising two stages: pre-training and fine-tuning. In the pre-training stage, the framework extracts 2D keypoint sequences from diverse motion data sources and applies random masking and noise to them. Subsequently, a motion encoder is trained to recover 3D motion from the corrupted 2D keypoints. This proxy task requires the motion encoder to infer the underlying 3D human structure from temporal motion and restore missing and erroneous data, thus implicitly learning common knowledge of human motion, such as joint topology, physiological constraints, and temporal dynamics. The authors introduced a dual stream spatial temporal transformer (DSTformer\cite{zhang2024dstformer}) as the motion encoder to capture long-range dependencies among skeletal keypoints. They hypothesized that motion representations learned from large-scale and diverse data can be shared across different downstream tasks, enhancing their performance. Therefore, for each downstream task, only fine-tuning of the pre-trained motion representations and a simple regression head network are required. Peng et al. proposed a 3D human body generation method based on neural radiation fields (NeRF) \cite{mildenhall2021nerf}. This method represents the geometry and texture of the human body using implicit neural networks, generating highly realistic 3D human models from different viewpoints. It can also adapt the appearance of the human body based on input pose information. In mesh reconstruction, the normal vector for each point on the model is computed. We will utilize the normal vector computation code in Neural body to obtain the normal vectors.
}

\section{Dataset}
We used Human3.6M \cite{ionescu2013human3} dataset with the train and test splits to do experiments. Human3.6M includes 15 different types of actions such as walking, running, and calling, which provide a comprehensive data foundation for MDC-Net. Human3.6M features 32 human keypoints, but we did not use all of them. Instead, we removed some less important points and used 16 remaining keypoints to construct a human body for training. Following previous works \cite{chen2023humanmac, martinez2017human}, we use subjects S1, S5, S6, S7 and S8 for training and S9, S11 for testing. 

\section{Method}

\subsection{Process Flow}
 This section will provide a detailed explanation of our methodology. As shown in Fig.~\ref{fig1}, we divide a human motion sequence (total M frames) into three parts: $H1$, $P$, and $H2$. The $H1$ sequence consists of the frames of human motion 1, while the $H2$ sequence is composed of the frames of human motion 2. The completion sequence P is what we need to generate. We applied different padding strategies to prediction part. We padded it by using the last frame of human motion 1 and the first frame of human motion 2, zero matrices, the last frame of motion 1 and the first frame of motion 2, shown in Fig.~\ref{fig3}. As shown in the figure, we sample the last x frames of $H1$, represented as $h1$ in the figure, and the first k frames of $H2$, represented as $h2$. These two, along with p, form the input as $h1 + P + h2$, which is then fed into the model as a whole. Based on our experiments, splitting the filling equally between $H1$ and $H2$ yields the best results. The operations described above can be easily implemented using torch's append and split functions.

\begin{figure}[H]
\centering
\includegraphics[width=1.0\linewidth]{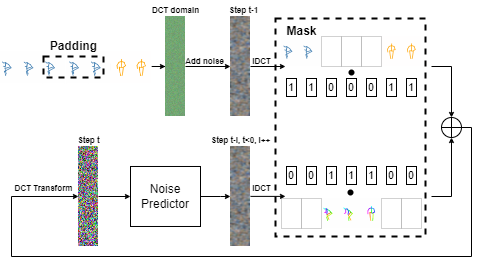}
\caption{This is the flowchart of MDC-Net. We embed the input data into the DCT domain and use a mask to get our required part of these sequences.}
\label{fig2}
\end{figure}

\begin{figure}[H]
\centering
\includegraphics[width=0.8\linewidth]{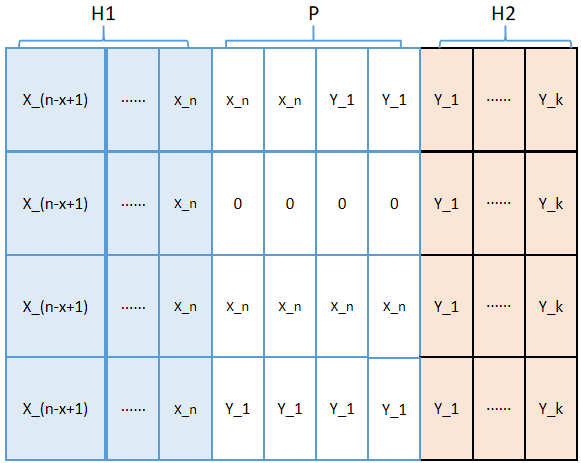}
\caption{\B{Different padding strategies.} We conducted experiments on P using the following four strategies: From first line to fourth line of figure, 1. Filling P with the last frame of $H1$ and the first frame of $H2$ respectively; 2. Setting all element of P to zero. 3. Filling all elements of P with the last frame of $H1$; 4. Filling all elements of P with the first frame of $H2$.}
\label{fig3}
\end{figure}

\begin{figure}[H]
\centering
\includegraphics[width=1.0\linewidth]{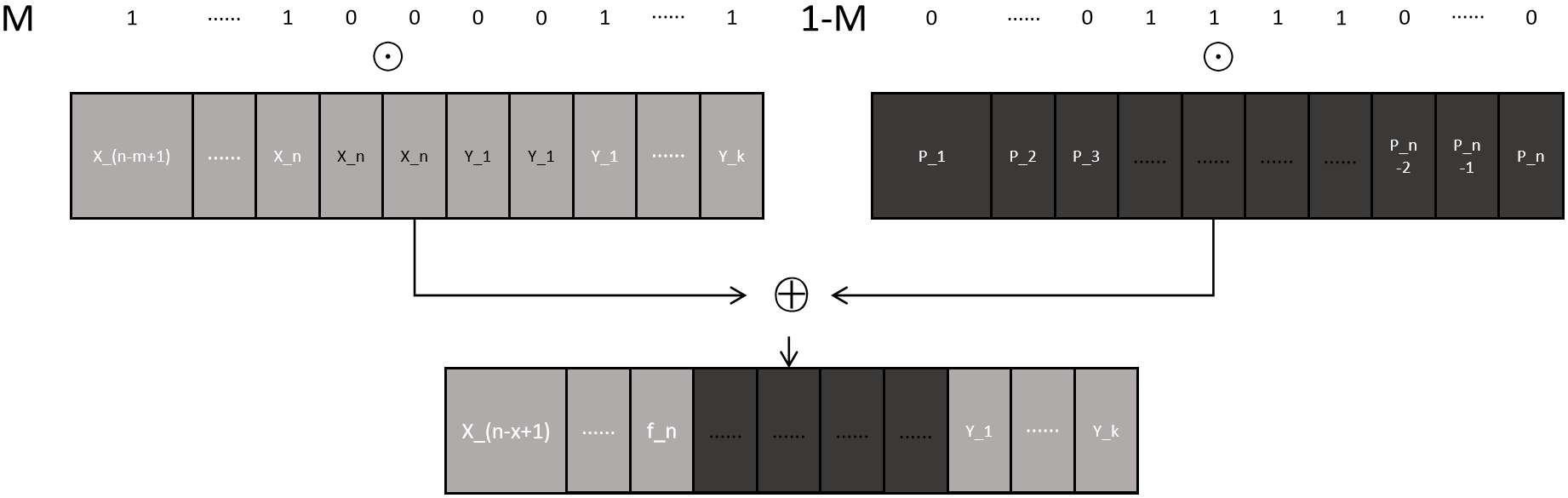}
\caption{Mask.The gray segment represents the sequences after padding, while the black segment represents the noise sequence $P$. $H1\{ X_(n-m+1), ... , X_n\}$ and $H2\{ Y_1, ... , Y_k \}$ are the motion sequence that input into the model. By multiplying the matrix M with the gray sequences, the inital motion sequences can be extracted. Then, by multiplying the 1-M with the black sequence, the sequence that need to be generted can be extracted. Finally, adding these two parts togther yields the complete sequence.}
\label{fig4}
\end{figure}

As shown in Fig.~\ref{fig2}, before adding noise, we transform human motion sequences from the time domain to the frequency domain using DCT. Previous works \cite{chen2023humanmac} and \cite{huang2022winnet} adapt this technology, which let it increase its performance better. Adding noise up to step t-1, we perform iDCT transformation to convert the frequency-domain signal back to the time-domain signal. At the same time, we also pass pure noise through our noise model, then perform a denoising process to obtain the frequency domain signal at step t-1, followed by an iDCT transformation to convert it back to the time domain signal. As mentioned in Section 2.1, we can predict the prediction part only using the last few frames of $H1$ and the initial few frames of $H2$. In practice, the number of frames taken from $H1$ and $H2$ can be different. As shown in Fig.~\ref{fig3}, we take the last x frames of $H1$ and the first k frames of $H2$. For motion completion, we utilize masking techniques, as shown in Fig.~\ref{fig4}, we use a matrix composed of 0 and 1 to remove the human motion sequences that we do not need and keep the sequences that will be used for training. The result after masking is given by:

\begin{equation}
y_{t-1}=M \cdot iDCT\left(y_{t-1}^{a}\right)+(1-M) \cdot iDCT\left(y_{t-i}^{d}\right)
\end{equation}

Here, $y_{t-1}^{a}$ denotes the sequence after denoising while $y_{t-1}^{d}$ denotes the sequence after adding noise.

\subsection{Motion Diffusion Completion Network}

\begin{figure}[H]
\centering
\includegraphics[width=1.0\linewidth]{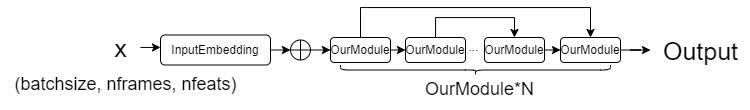}
\caption{Baseline. In the figure, nframes represents the \B{total} n frames that input into model. Similarly, nfeats represents the number of keypoints and their xyz coordinates. }
\label{fig5}
\end{figure}

Structure: The structure of MDC-Net is shown in Fig.~\ref{fig5}. 
The structure of our modules are connected one by one while the paired modules are connected via the skip connections where we use the skip connection from \cite{he2016deep}.
The number of our modules is $N$. 
This structure resembles the structure used in HumanMAC \cite{chen2023humanmac}. However,
HumanMAC uses the number of eight modules while we use the number of four modules.

\begin{figure}[h]
\centering
\includegraphics[width=1.0\linewidth]{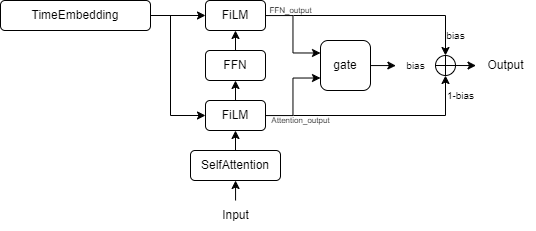}
\caption{In our module, we introduced a gate structure into a normal transform input embedding, performing a weighted sum of attention result and ffn result.}
\label{fig6}
\end{figure}

\textbf{Gate module.} We introduce a gate module, which consists of a linear layer followed by a sigmoid function, to calculate the bias. As shown in Fig.~\ref{fig6}, the bias output of the gate module determines about which features contribute to the final output. The final output is given by:

\begin{equation}
y_{t-1}=\mbox{bias} \cdot \mbox{FFNoutput} + (1- \mbox{bias}) \cdot \mbox{AttentionOutput}
\end{equation}

The gate module connects the self-attention mechanism and the FFN layer. Self-attention excels at capturing global contextual information, while the FFN network specializes in capturing local and high-level features. Using the weighted sum, these two types of features can be integrated, providing the model with a more comprehensive understanding of the input information.

\textbf{TimeEmbedding.} To effectively model temporal dependencies in sequential data, we use time embeddings in our framework. The embedding explicitly encodes the temporal sequence of the input data, facilitating better temporal representation. We introduce a Position TimeEmbedding Module. By incorporating information from different time scales, the generated motions may become smoother and more natural.

\section{Result and Discussion}
\subsection{Implementation} 
\textbf{Details.} This study trained for a total of 1000 epochs. For the diffusion model, there are 1000 noise addition processes. The trajectory sequence length is set to 125 frames, with the first 10 frames as the history part, the middle 90 frames as the prediction part, and the last 15 frames as the future part. Using Adam as the optimizer strategy, the learning rate is set to 0.0003.

\textbf{Evaluation.} We use the metrics APD, ADE, FDE, MMFDE, MMADE that established by \cite{chen2023humanmac} for our evalution.

\textbf{Environment.} All of the experiments are implemented in a GEFORCE RTX 3060 12G, Ubuntu 20.04.

\subsection{Quantitative Analysis}

We compare MDC-Net with HumanMAC and MDM \cite{mdm}. The results are provided in Table 1. As can be seen, although MDC-Net does not achieve a comprehensive lead, it exhibits better results in the ADE, the FDE, and the MMADE metrics. The average pairwise distance (APD) is the L2 distance between all motion examples, used to measure the diversity of results. The average displacement error (ADE) is the smallest average L2 distance between the ground truth and the predicted motion, indicating the accuracy over the entire sequence. The final displacement error (FDE) is the L2 distance between the predicted result and the ground truth in the last prediction frame. The multimodal-ADE (MMADE) is the multimodal version of the ADE metric, where future motions in the ground truth are grouped based on similar observations. The multimodal-FDE (MMFDE) is the multimodal version of the FDE metric, where multiple future predictions are grouped by similar observations. In this case, the error is calculated in the last prediction frame\cite{chen2023humanmac}. 

\begin{table}[H]
\caption{\B{Experimental results on different models. Bolded numbers denote the better results. Average Pairwise Distance (APD): The L2 distance between all motion examples, used to measure the diversity of results. Average Displacement Error (ADE): The smallest average L2 distance between the ground truth and predicted motion, indicating the accuracy of the entire sequence. Final Displacement Error (FDE): The L2 distance between the predicted result and the ground truth in the last prediction frame. Multi-Modal-ADE (MMADE): The multi-modal version of ADE, where future motions in the ground truth are grouped based on similar observations. Multi-Modal-FDE (MMFDE): The multi-modal version of FDE, where multiple future predictions are grouped by similar observations, and the error is calculated at the last prediction frame}}
\centering
\begin{tabular}{@{}lllll@{}}
\toprule
\multicolumn{5}{c}{Human3.6M}                \\ \midrule
         & ADE \( \downarrow \)    & FDE \( \downarrow \)    & MMADE \( \downarrow \)  & MMFDE \( \downarrow \)  \\
MDC-Net  & \textbf{0.2195} & \textbf{0.0769} & \textbf{0.5716} & 0.8077 \\
MDM      & 0.3526 & 0.1331 & 0.6383 & \textbf{0.7276} \\
HumanMAC & 0.2352 & 0.0839 & 0.5718 & 0.7946 \\ \bottomrule
\end{tabular}
\end{table}

\subsection{Ablation Study}
We conduct ablation experiments on MDC-Net, including the structure of our prediction network; different diffusion variance noise strategies; the settings of our module.

\textbf{Structure of our prediction network.} We tested the performance of MDC-Net with different numbers of layers. In Table 2, a performance comparison is presented between our model and the HumanMac model. We set our skip connection structure into 4 and 8 layers. The 4-layer model achieved much better results than the 8-layer model in ADE, FDE, MMADE, MMDFE and the model size. The 4-layer model outperforms the 8-layer model in terms of ADE, FDE, MMADE, and MMFDE, while maintaining a smaller parameter size of only 16.84M. In contrast, the 8-layer model achieves a significantly higher APD, indicating increased diversity in the generated motion sequences. However, its ADE and FDE errors increase considerably. At the same time, the parameter size of the 8-layer model reaches 38.90M, leading to a substantial increase in computational cost.

\begin{table}[H]
\centering
\caption{Comparison of 4-layer and 8-layer models of our MDC-Net on the Human3.6M dataset.}
\resizebox{\textwidth}{!}{ 
\begin{tabular}{@{}l c c c c c c c@{}}
\toprule
\multicolumn{8}{c}{Human3.6M} \\ \midrule
\textit{Model} & Layers & APD \( \uparrow \) & ADE \( \downarrow \) & FDE \( \downarrow \) & MMADE \( \downarrow \) & MMFDE \( \downarrow \) & Size \\ \midrule
MDC-Net 4  & 4  & 3.1029 & \textbf{0.2195} & \textbf{0.0769} & \textbf{0.5716} & \textbf{0.8077} & \textbf{16.84}M \\
MDC-Net 8  & 8  & \textbf{6.0502} & 0.5544 & 0.3730 & 0.7964 & 0.8217 & 38.90M \\ 
HumanMac 8 & 8  & 3.3563 & 0.2352 & 0.0839 & 0.5718 & 0.7946 & 28.40M \\ \bottomrule
\end{tabular}
}
\label{tab:layers_results}
\end{table}

\begin{figure}[h]
\centering
\includegraphics[width=0.6\linewidth]{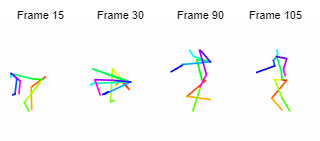}
\caption{Visualizaiton results of Sqrt strategy. This figure shows the experimental results after applying the sqrt strategy. Frame 15, frame 30, frame 90 and frame 105 are sampled from the generated completion motion. These frames show the transition process from $H1$ to $H2$, and exhibit issues such as distortion, causing the motion to completely violate the physical laws of the human body.}
\label{fig7}
\end{figure}

\textbf{Different diffusion variance noise scheduling.} We conduct different diffusion variance noise strategies including the sqrt, the sigmoid, the linear, and the cosine sampling strategies for quantitative experiments. In Table 3, we compare different noise scheduling strategies when training the model. The sigmoid strategy performs best in MMFDE, while the sqrt strategy has the highest APD, meaning it creates more diverse motions. However, the cosine strategy achieves the best results in ADE, FDE, and MMADE, making it a more balanced choice. Although the sqrt strategy increases diversity, as shown by its high APD, it does not perform well in visualizations, as shown in Fig.~\ref{fig7}. Since our task focuses on generating smooth and natural motion transitions rather than maximizing diversity, we prioritize logical motion flow from human motion 1 to human motion 2.

\begin{table}[H]
\caption{\B{Performance comparison of different noise scheduling strategies. The cosine strategy achieves the best results in ADE, FDE, and MMADE, while the sqrt strategy excels in APD, indicating higher diversity. However, the cosine strategy provides a more balanced performance suitable for smooth motion transitions.}}
\centering
\begin{tabular}{@{}lccccc@{}}
\toprule
\multicolumn{6}{c}{Human3.6M} \\ \midrule
\textit{Strategies} & APD \( \uparrow \) & ADE \( \downarrow \) & FDE \( \downarrow \) & MMADE \( \downarrow \) & MMFDE \( \downarrow \) \\ \midrule
Cosine & 3.1029 & \textbf{0.2195} & \textbf{0.0769} & \textbf{0.5716} & 0.8077 \\
Linear & 2.9405 & 0.3357 & 0.1207 & 0.6583 & 0.7700 \\
Sigmoid & 3.0875 & 0.3368 & 0.1243 & 0.6501 & \textbf{0.7677} \\
Sqrt & \textbf{6.0502} & 0.5544 & 0.3730 & 0.7964 & 0.8217 \\ \bottomrule
\end{tabular}
\label{tab:scheduler_results}
\end{table}

\textbf{Settings of our module.} To better understand the contribution of each component within MDC-Net, we designed the ablation experiments as follows:
\begin{itemize}
  \item [1)] 
  Removing the gate module: We evaluated the performance without the gate module.     
  \item [2)]
Removing the multiscale time module: We evaluated the performance without the multiscale time module. 
  \item [3)]
  Removing the gate and the multiscale time modules: We evaluated the performance without the gate module and the multiscale time module.
\end{itemize}
In Table 4, the gate module and the multiscale time module contribute significantly to the performance of the model. Among them, the gate module achieved significant improvements in ADE and FDE. Summing the feature outputs of the self-attention and the FNN modules, it makes the generation more accurate.

\begin{table}[H]
\caption{\B{This table shows the results of ablation experiments on the Human3.6M dataset,} comparing the performance on different settings of the modules. Baseline setting is the simplest module stettings. +OurTimeEmbedding is added the time embedding module. +GateModule is added the gate module. The bold numbers indicate the better results compared to the baseline.}
\centering
\resizebox{\textwidth}{!}{%
\begin{tabular}{@{}lllllll@{}}
\toprule
\multicolumn{7}{c}{Human3.6M} \\ \midrule
\textit{No.} & Model & APD \( \uparrow \) & ADE \( \downarrow \) & FDE \( \downarrow \) & MMADE \( \downarrow \) & MMFDE \( \downarrow \) \\ \midrule
1 & Baseline & 3.3563 & 0.2352 & 0.0839 & 0.5718 & 0.7946 \\
2 & +OurTimeEmbedding & \textbf{3.3941} & 0.2355 & 0.0848 & \textbf{0.5705} & \textbf{0.7932} \\
3 & +GateModule & 3.0654 & \textbf{0.2195} & \textbf{0.0769} & 0.5727 & 0.8082 \\
4 & +OurTimeEmbedding+GateModule & \textbf{3.1029} & \textbf{0.2176} & \textbf{0.0767} & \textbf{0.5716} & 0.8077 \\ \bottomrule
\end{tabular}%
}
\label{tab:human3.6m}
\end{table}

\subsection{Visualization Results}
\B{In this section, we compare the visualization using HumanMac and those using MDC-Net.
Then, we show the various motion transitions which are generated by MDC-Net.}

\begin{figure}[H]
\centering
\includegraphics[width=0.8\linewidth]{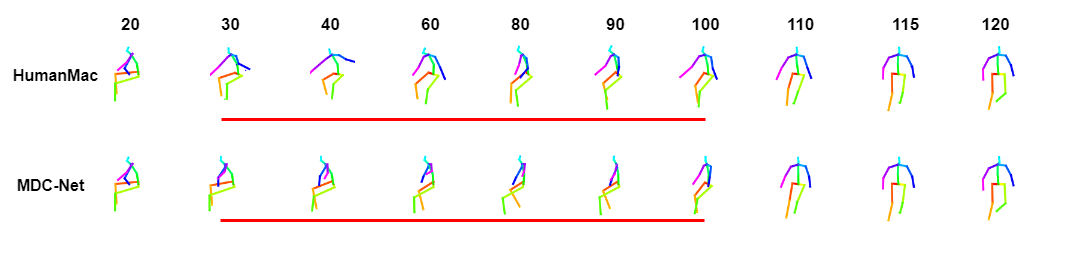}
\caption{\B{Comparisons using HumanMAC and using MDC-Net. These figures show the visualization results of the transition from sitting to walking using HumanMAC and those using MDC-Net}. There are a total of 125 frames. We sampled images from frames 20, 30, 40, 60, 80, 90, 100, 115, and 120 for display. The images demonstrate that using MDC-Net, the body proportions in the completion motion remain more normal, and the transition process is smoother.}
\label{fig8}
\end{figure}

We conducted an experiment on the transition from "Sitting" to "Walking", using the last 25 frames of "Sitting" and the first 10 frames of "Walking" as input to the model, with 90 frames for motion completion. The experimental results were used to create Fig.~\ref{fig8}. As shown in the red underline, using HumanMAC, the human's torso becomes noticeably deformed and disproportionate. Additionally, from frame 30 to frame 100, the turning motion changes too quickly and lacks smoothness. In contrast, with MDC-Net, the motion is more natural, and the human's torso maintains proper proportions.

\begin{figure}[H]
\centering
\includegraphics[width=0.85\linewidth]{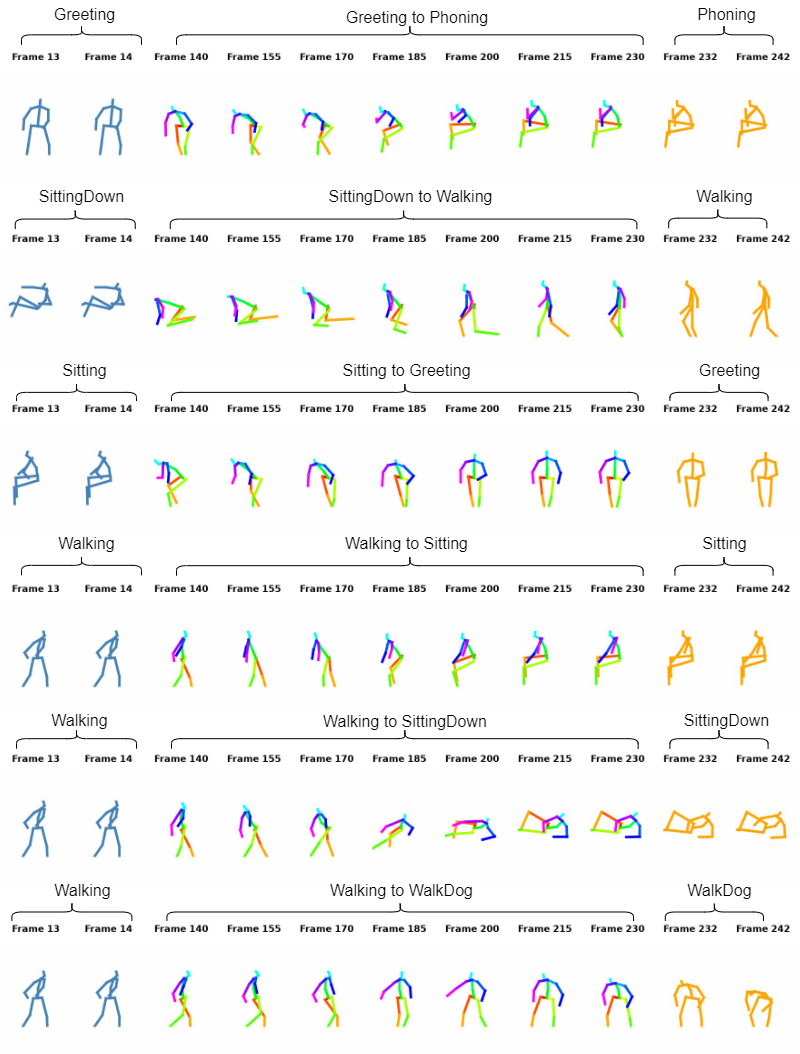}
\caption{Visualization results of human motion completion. The actions on the left represented by the blue-colored human skeletion is the visualization of $H1$. \B{We sampled the frames 13 and 14. The actions on the right represented by the orange-colored human skeletion is the visualization of $H2$. We sampled the frames from 232 to 242. All $H1$ and $H2$ are randomly sampled from the Human3.6M dataset. The middle part, the colorful human skeletons denote the motion completion, showing the transformation process from $H1$ to $H2$.}}
\label{fig9}
\end{figure}

For visualization, we choose six human actions: Greeting, Phoning, SittingDown, Walking, Sitting, WalkDog. As shown in Fig.~\ref{fig9}, we show six cases of motion completion: GreetingToPhoning, SittingDownToWalking, SittingToGreeting, WaitingToSitting, WalkingtoSittingDown, Walking to WalkDog. \B{Since the total number of frames is too large in order to display the all sequence, we show only the last two frames of $H1$ and the first two frames of $H2$ among them.} The frames were sampled every 15 frames for the motion completion.

\section{AI Motion Capture} 
\B{This section explains how our MDC-Net is deployed to the AI motion capture, focusing on how new motion sequences can be generated from the fixed observations to make motion capture more flexible and adaptable. Traditional AI motion capture uses video recordings. This means that it can only copy existing movements and cannot create new ones. This is a problem because every motion must be predefined. This situation makes it hard to use in situations such as virtual characters and games where new movements are needed. In order to resolve this problem, we introduce a diffusion-based motion completion method that creates more diverse and interesting human motion sequences by combining a few discrete human motions. For example, a motion sequence like this can be generated: a person walking, sitting down at a certain spot, then getting up and walking to the bedside, and finally lying down. We also propose a method for obtaining IMU data from these human motion sequences. Traditional IMU data collection usually requires the use of specialized motion capture equipment, whereas our method enables the rapid and cost-effective acquisition of large amounts of IMU data.}

\subsection{Human Motion Completion}
\begin{figure}[H]
\centering
\includegraphics[width=0.5\linewidth]{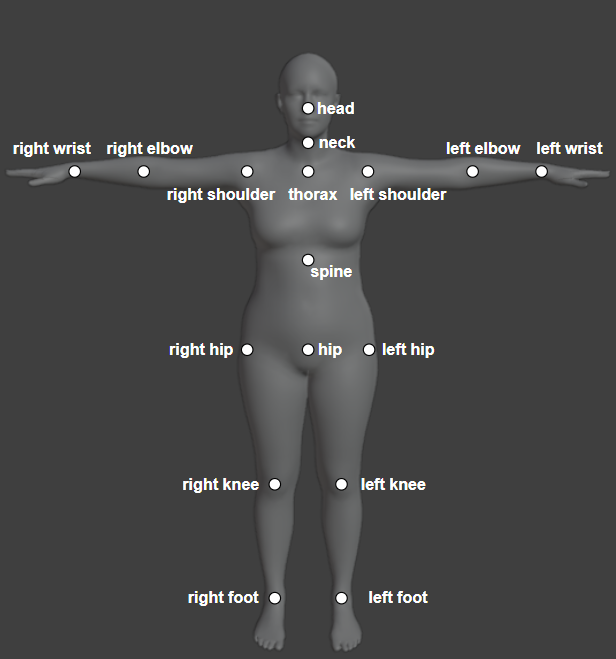}
\caption{Virtual human structure. It consists 17 joints}
\label{fig10}
\end{figure}

First, we select the "Greeting" and "Phoning" actions from the Human3.6M dataset. Then, we applied our motion completion technique to these actions. It is important to note that we did not input the full sequences of these actions into MDC-Net. Instead, we selected the last 15 frames of the "Greeting" action and the first 20 frames of the "Phoning" action, completing an additional 90 frames for each. This is illustrated in Table 5. We call a generated action sequence the "GreetingToPhoning" action sequence. The choice of 15 and 20 frames is based on experimental considerations. In terms of the structure of human skeleton, we adopted a human skeleton structure consisting of 17 joints, as illustrated in Fig.~\ref{fig10}.

\subsection{\B{Mesh Estimation} from Human Skeleton}
\B{3D human pose and \B{mesh estimation} aims to recover 3D locations of human joints and mesh vertex simultaneously\cite{choi2020pose2mesh}.} In the mesh estimation section, we adopted MotionBERT \cite{motionbert}. The original MotionBERT workflow involves using AlphaPose \cite{Alphapose} to predict the 2D coordinates of human joints, followed by depth estimation to obtain the (x, y, z) coordinates of each joint. However, we did not follow this process. Since we already have the 3D coordinates of each joint, we bypassed the joint detection and depth estimation steps, directly entering our 3D data into the MotionBERT model to estimate the parameters of the SMPL model. In this way, we obtained the SMPL 3D mesh model for each frame of our completion action and, most importantly, extracted the normal vector information for each vertex of the mesh model.

\subsection{Sensor Data from Mesh Model}
\begin{figure}[H]
    \centering
    \includegraphics[width=0.8\linewidth]{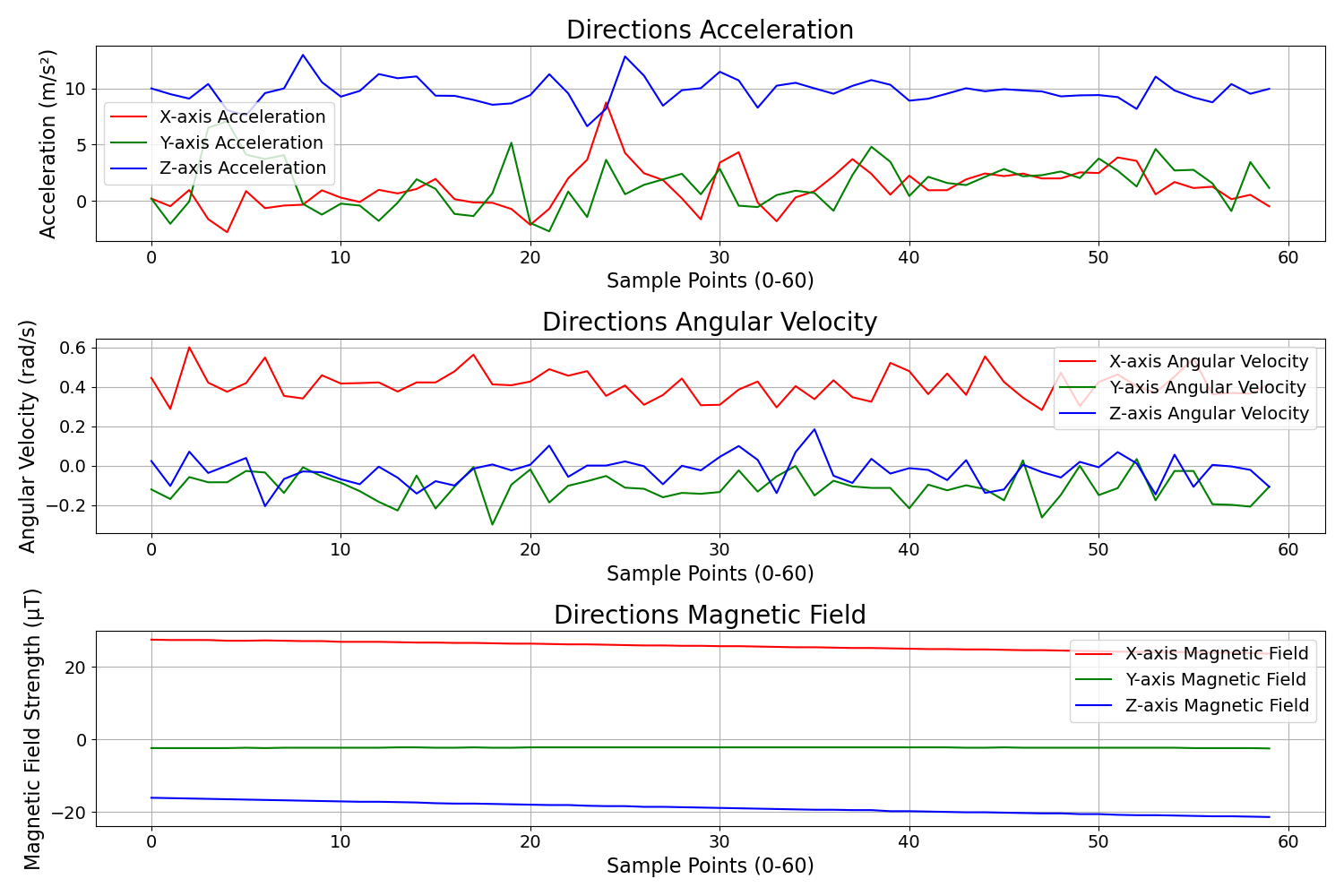}
    \caption{This figure describes a person giving directions. The first row of the figure shows the acceleration changes, while the second row shows the angular velocity. During the process, the person irregularly raises and waves their left wrist, resulting in chaotic waveforms.}
    \label{fig11}
\end{figure}

\begin{figure}[H]
    \centering
    \includegraphics[width=0.8\linewidth]{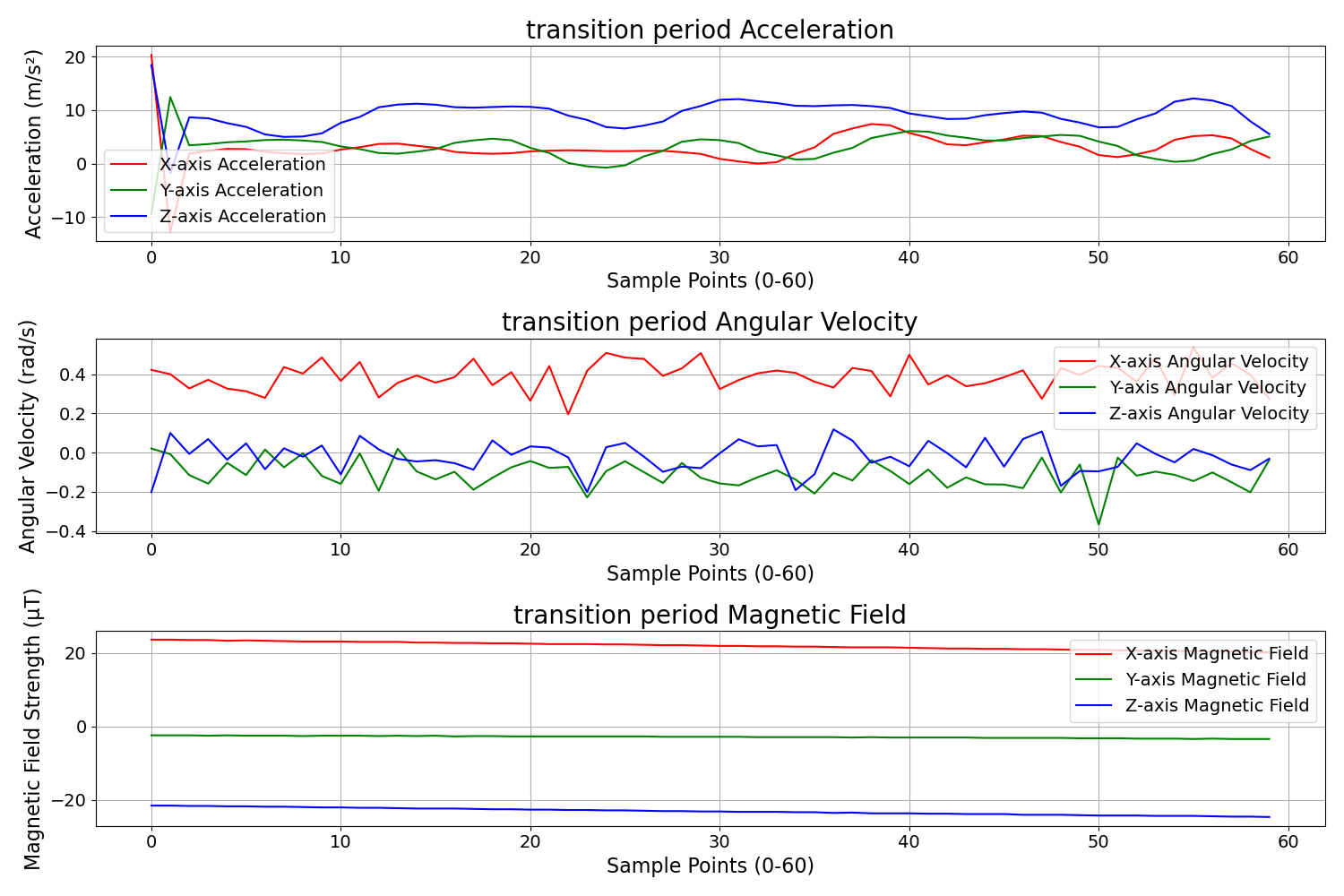}
    \caption{This figure describes the process that the transition from directions to photo. This person places their hands in front of him and take a photo. In the first row of figure, the acceleration curve shows a period of intense fluctuation at the beginning, reflecting the rapid motion of the hands being brought back to the front, which causes a significant change in acceleration. Then hands stop in front of the body. Throughout the process, the angular velocity of the left wrist changes very little, staying close to a steady value with slight swings.}
    \label{fig12}
\end{figure}

\begin{figure}[H]
    \centering
    \includegraphics[width=0.8\linewidth]{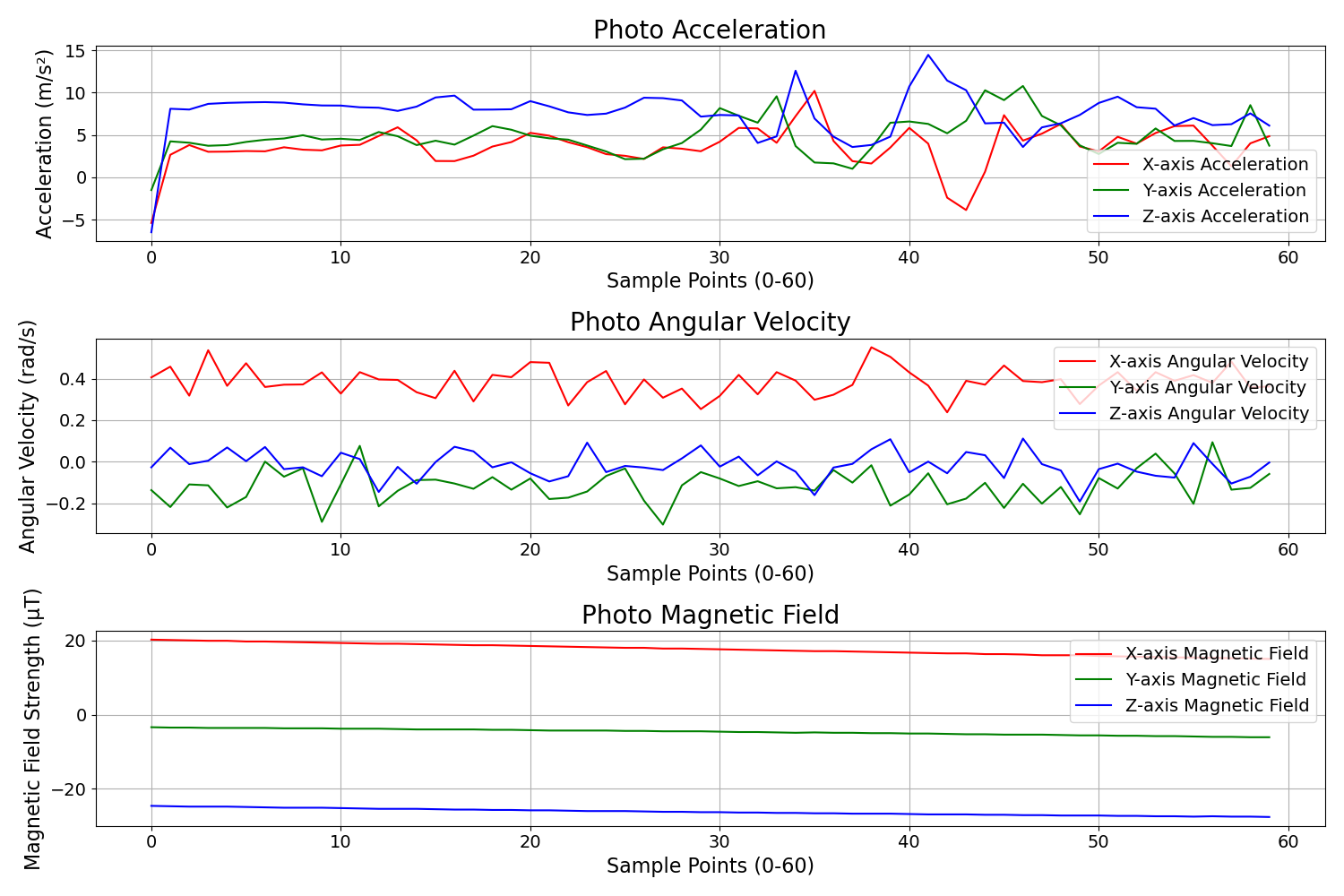}
    \caption{This figure describes the process a person is taking a photo of one location and then take a photo of others. This causes the acceleration waveform to remain stable for a while, then become chaotic.}
    \label{fig13}
\end{figure}

\begin{figure}[H]
    \centering
    \includegraphics[width=0.8\linewidth]{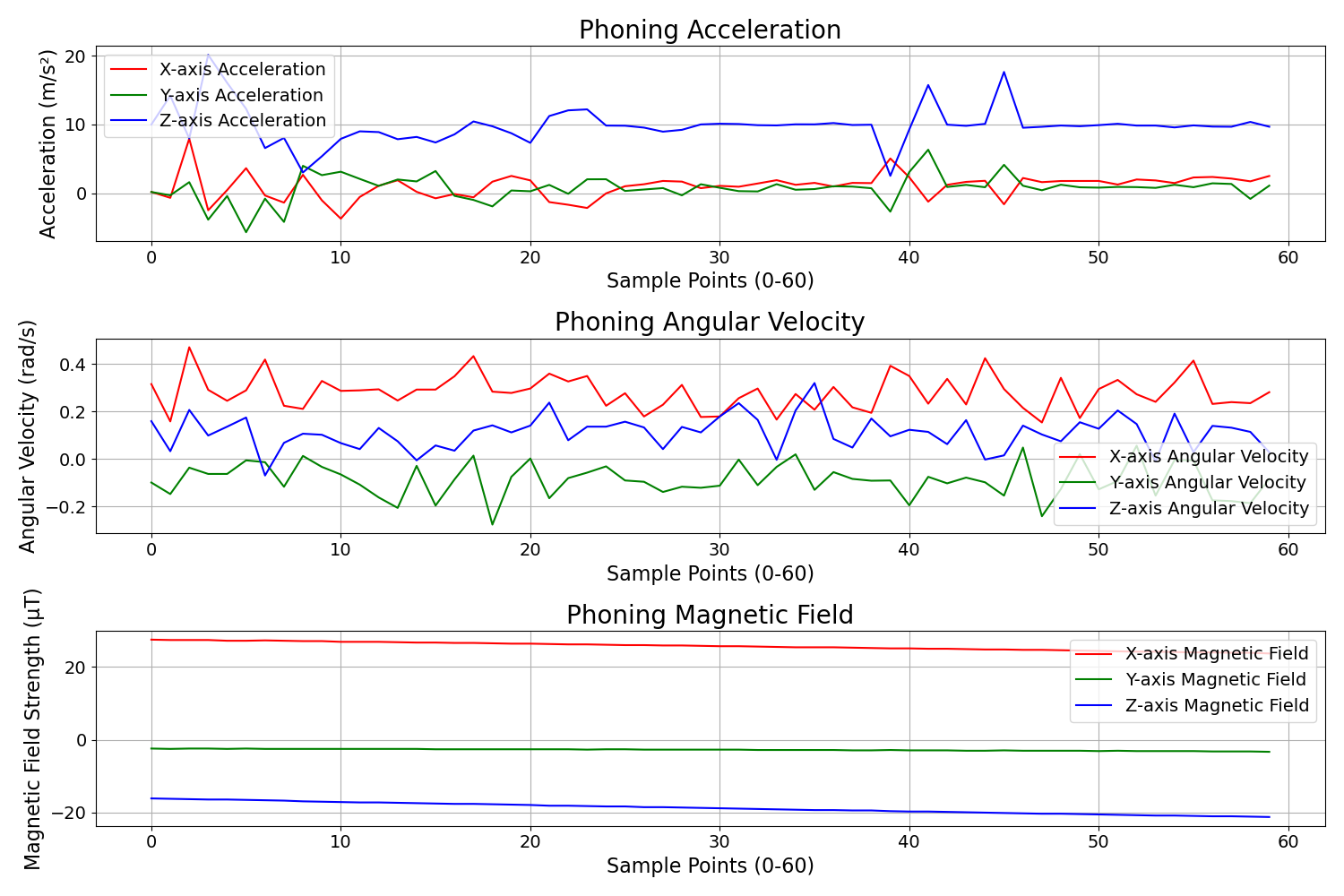}
    \caption{This figure describes the process that a person is sitting and talking on the phone. He is holding the phone in his left hand, resting it against his ear.}
    \label{fig14}
\end{figure}

\begin{figure}[H]
    \centering
    \includegraphics[width=0.8\linewidth]{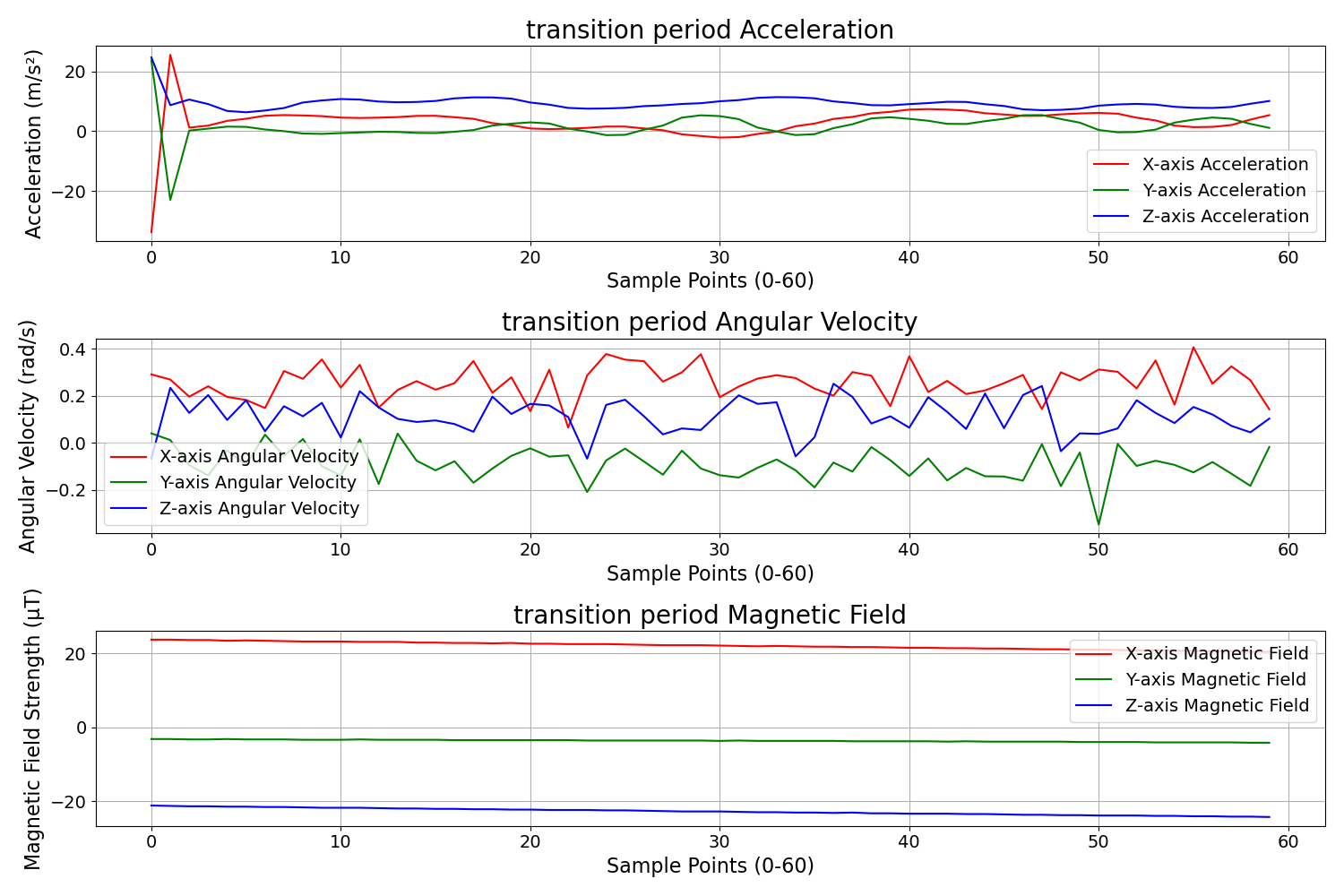}
    \caption{This figure describes the process that the transition from phoning to walking. This person lowers his left hand and at the same time stands up and start walking.}
    \label{fig15}
\end{figure}

\begin{figure}[H]
    \centering
    \includegraphics[width=0.8\linewidth]{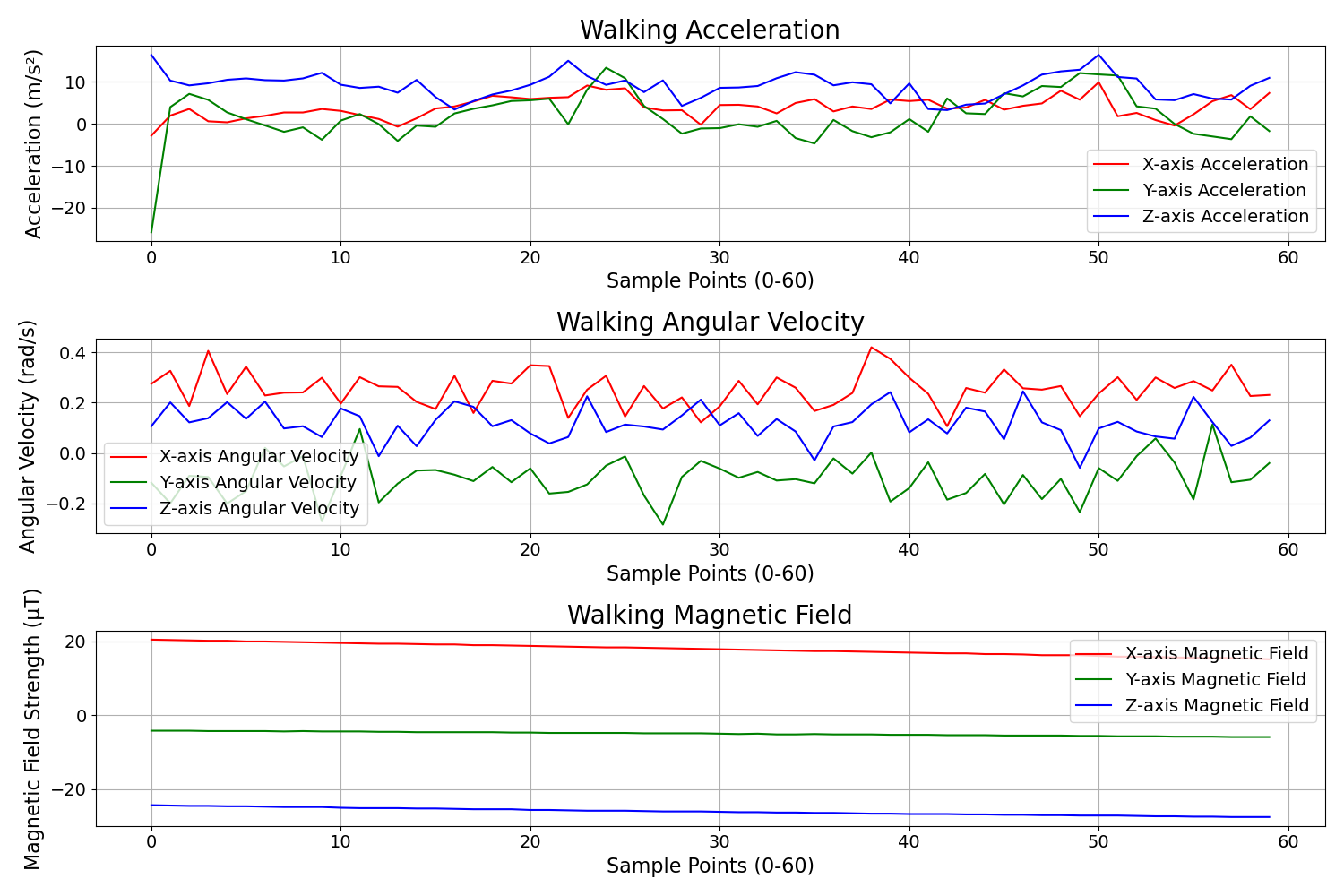}
    \caption{This figure describes the acceleration and angular velocity of a person's left wrist while walking.}
    \label{fig16}
\end{figure}

To obtain the sensor data for the left wrist, we first needed to determine its position in the model. We imported the SMPL \cite{loper2023smpl} mesh model file into Blender and switched to edit mode, allowing us to view the position of each vertex along with its corresponding index. In this context, the vertex index corresponds to the position in the SMPL \cite{loper2023smpl} vertex array. We selected the vertex with index 2208 as the location for the left wrist, which was an experimental choice. Once we identified the vertex of the left wrist, we were able to compute the normal vector using the mesh model for each frame, as mentioned in the previous section. We utilized the NeuralBody \cite{peng2021neural} normal vector computation code, which allowed us to obtain the normal vector for the left wrist. At this point, we had both the normal vector and the 3D coordinates for the left wrist (since our generative model directly outputs the 3D coordinates for each node, no additional calculation for the coordinates of the left wrist was necessary). We then input the normal vector and coordinate data into the MATLAB IMU module to fit the sensor data and generate the plot shown in Fig.~\ref{fig11}--\ref{fig16}. We sampled sixty points. The plot illustrates the acceleration and angular velocity of the left wrist when a person does actions from phoning to walking and from phoning to walking. Since we did not plot the magnetic field, this line is a straight line.

\begin{table}[htbp]
\caption{
We prepare two original actions from Human3.8M: greeting and phoning. Each action has a total of 125 points at $50$Hz. The inputs to our diffusion model are the last 15 points of greeting and the first 20 points of phoning. The diffusion model generates points between these two actions, producing 90 points. Therefore, the resulting generated points consist of 125 points (15 points from greeting, 90 points for the transition from greeting to phoning, and 20 points from phoning).
}
\centering
\resizebox{0.9\linewidth}{!}{  
\begin{tabular}{|c|c|c|c|c|}
\hline
         & Original points & Used points & Generated points & Total points \\ \hline
Greeting & 125             & 15          & 0                & 125          \\ \hline
Phoning  & 125             & 20          & 0                & 125          \\ \hline
GreetingToPhoning & 340   & 35          & 90               & 340          \\ \hline
\end{tabular}
}
\label{tab:points}
\end{table}

\section{Conclusion}
\B{We propose MDC-Net, a model capable of handling input motion sequences of any length and generating output motion sequences of any length. We demonstrate that MDC-Net operates with lower memory usage and computational complexity compared to HumanMAC. Additionally, we show that MDC-Net can be deployed to generate virtual IMU data at specific joints from human motion sequences.}
%
MDC-Net focuses on generating missing action sequences between fragmented human motions, enabling the creation of long and coherent motion sequences. By incorporating a gate module and a position-time embedding module, MDC-Net achieves competitive results on the Human3.6M dataset. Specifically, MDC-Net outperforms existing methods such as FlowMDM and HumanMAC in terms of ADE, FDE, and MMADE metrics, while maintaining a smaller model size of 16.84M compared to HumanMAC’s 28.40M.
\B{Additionally, we propose a method to obtain sensor data for specific body parts from generated human motions. This approach eliminates the need for specialized hardware, reducing costs and providing substantial data support for AI-driven motion capture.}

{\sf Limitations and future work.} 
Our approach has certain limitations. In some cases, the generated transitions deviate from realistic human movement patterns and physical laws. For example, during a transition from sitting to walking, unnatural motions may occur, such as the legs extending downward instead of the upper body rising first. Additionally, converting from a human skeleton to a mesh model may introduce inaccuracies, leading to larger errors in angular velocity.
Future work will focus on incorporating real-world physical constraints and biomechanical principles into the generation process to enhance realism and ensure physically plausible transitions.

\bibliographystyle{plain}
\bibliography{bibtex}

\end{document}